\documentclass{article} 
\usepackage{nips14submit_e,times}
\usepackage{hyperref}
\usepackage{url}
\usepackage{amsmath,amsfonts,amssymb}
\usepackage{graphicx}
\usepackage{algorithm}
\usepackage{algorithmic}
\usepackage{caption}
\usepackage{subcaption}
\usepackage{bbm}
\usepackage[numbers]{natbib}
\usepackage{dsfont}
\usepackage{array}

\title{Training for Fast Sequential Prediction Using Dynamic Feature Selection}

\author{
Emma Strubell \qquad Luke Vilnis \qquad Andrew McCallum\\
School of Computer Science\\
University of Massachusetts, Amherst\\
Amherst, MA 01002 \\
\texttt{\{strubell, luke, mccallum\}@cs.umass.edu}
}

\DeclareMathOperator*{\argmax}{arg\,max}

\nipsfinalcopy 

\newcommand{\comment}[1]{}

\newcommand{\newcite}[1]{\citet{#1}}

\begin{document}

\maketitle

\begin{abstract}
We present paired learning and inference algorithms for significantly
reducing computation and increasing speed of the vector dot products
in the classifiers that are at the heart of many NLP components.  This
is accomplished by partitioning the features into a sequence of
templates which are ordered such that high confidence can often be
reached using only a small fraction of all features.  Parameter
estimation is arranged to maximize accuracy and early confidence in
this sequence. We present experiments in
left-to-right part-of-speech tagging on WSJ, demonstrating that we can
preserve accuracy above 97\% with over a five-fold reduction in
run-time.
\end{abstract}

\section{Introduction}


The heart of the prediction computation in linear NLP models is
a dot-product between a dense parameter vector and a sparse feature
vector. The bottleneck in these models is then often a combination of
 numerical operations and potentially expensive feature extraction.
  However, in many cases not all features are necessary
for accurate prediction.

We present a simple yet novel approach to improve processing speed by
dynamically determining on a per-instance basis how many features are
necessary for a high-confidence prediction. Our features are divided
into a set of \emph{feature templates}, such as {\sf\small
  current-token} or {\sf\small previous-tag} in the case of POS
tagging. At training time, we determine an ordering on the templates
such that we can approximate model scores at test time by
incrementally calculating the dot product in template ordering. We
then use a running confidence estimate for the label prediction to
determine how many terms of the sum to compute for a given instance,
and predict once confidence reaches a certain threshold.

We apply our method to left-to-right part-of-speech tagging in which we achieve accuracy above 97\% on the Penn Treebank WSJ corpus while running more than 5 times faster than our baseline.

In similar work, cascades of increasingly complex and high-recall models have been used for both structured and unstructured prediction. \newcite{viola2001rapid} use a cascade of boosted models to perform face detection. \newcite{weiss-taskar-10} add increasingly higher-order dependencies to a graphical model while filtering the output domain to maintain tractable inference. 

While most traditional cascades pass instances down to layers with increasingly higher recall, we use a single model and accumulate the scores from each additional template until a label is predicted with sufficient confidence, in a stagewise approximation of the full model score. Our technique applies to any linear classifier-based model over feature templates without changing the model structure or significantly decreasing prediction accuracy. 

Our work is also related to the field of learning and inference under
test-time budget constraints \cite{grubb-bagnell-12,trapznikov-saligrama-13}. 
However, common approaches to this problem
also employ auxiliary models to rank which feature to add next, and
are generally suited for problems where features are expensive to
compute ({\it e.g} vision) and the extra computation of an auxiliary
pruning-decision model is offset by substantial reduction in feature
computations \cite{weiss-taskar-13}. Our method uses confidence scores
directly from the model, and so requires no additional computation,
making it suitable for speeding up classifier-based NLP methods that
are already very fast and have relatively cheap features. In fact, the most 
attractive aspect of our approach is that it speeds up methods 
that are already among the fastest in NLP.


\newcite{he2012cost} have the same goal of speeding test time prediction by dynamically selecting features, but they also learn an additional model on top of a fixed base model, rather than using the training objective of the model itself. In the context of NLP, \newcite{he-et-al-13} describe a method for
dynamic feature template selection at test time in graph-based
dependency parsing using structured prediction cascades \cite{weiss-taskar-10}.
However, their technique is particular to the parsing
task---making a binary decision about whether to lock in edges in the
dependency graph at each stage, and enforcing parsing-specific,
hard-coded constraints on valid subsequent edges.  Furthermore, as
described above, they employ an auxiliary model to select features.


\section{Method}
\label{sec:alg}
We present paired learning and
inference procedures for feature-templated classifiers that optimize
both accuracy and inference speed, using a process of \emph{dynamic
  feature selection}. Since many decisions are easy to make in the
presence of strongly predictive features, we would like our model to
use fewer templates when it is more confident.  For a fixed, learned ordering
of feature templates, we build up a vector of class scores
incrementally over each prefix of the sequence of templates, which we
call the \emph{prefix scores}. Once we reach a stopping criterion
based on class confidence (margin), we discontinue computing prefix scores, and
predict the current highest scoring class. Our aim is to train
each prefix to be as good a classifier as possible without the
following templates, in order to minimize the number of templates
needed for accurate predictions. Template ordering is learned with
a greedy approach described at the end of this section.

\subsection{Definitions}

Our base classifier for sequential prediction tasks is a
\emph{linear model}. Given an input $x \in \mathcal{X}$, a set of
labels $\mathcal{Y}$, a feature map $\Phi(x,y)$, and a weight vector
$\mathbf{w}$, a linear model over \emph{feature templates} $\{\Phi_j(x,y)\}$ predicts the highest-scoring label as
\begin{align*}
y^* = \argmax_{y \in \mathcal{Y}} ~\mathbf{w} \cdot \Phi(x, y), ~~~~~~\text{where}~~~~~~ \mathbf{w} \cdot \Phi(x, y) = \sum_j \mathbf{w}_j \cdot \Phi_j(x,y).
\end{align*}
Our goal is to approximate the overall dot product score
sufficiently for purposes of prediction, while using as few terms of
the sum as possible.

Given a model that computes scores additively over template-specific scoring functions, parameters $\mathbf{w}$, and an observation $x \in {X}$, we can define the $i$'th \emph{prefix score} for label $y \in \mathcal{Y}$ as:
\begin{align*}
P_{i,y}(x, \mathbf{w}) = \sum_{j=1}^i \mathbf{w}_j \cdot \Phi_j(x,y),
\end{align*}
or $P_{i,y}$ when the choice of observations and weights is clear from context. Abusing notation we will also refer to the vector containing all $i$'th prefix scores for observation $x$ associated to each label in $\mathcal{Y}$ as $P_i(x, \mathbf{w})$, or $P_i$ when this is unambiguous.

Given a parameter $m > 0$, called the \emph{margin}, we define a function $h$ on prefix scores:
\begin{align*}
h(P_i, y) &= \max\{0,P_{i,y} + m - \max_{y'\neq y} P_{i,y'} \}
\end{align*}
This is the familiar structured hinge loss function as in structured support vector machines \cite{tsochantaridis2004support}, which has a minimum at $0$ if and only if class $y$ is ranked ahead of all other classes by at least $m$. Using this notation, the condition that some label $y$ be ranked first by a margin can be written as $h(P_i,y) = 0$.





\subsection{Inference}

At test time we compute prefixes until some label is ranked ahead of all other labels with a margin $m$, then predict with that label. At train time, we predict until the correct label is ranked ahead with margin $m$, and return the whole set of prefixes for use by the learning algorithm. If no prefix scores have a margin, then we predict with the final prefix score involving all the feature templates.

\subsection{Learning}
\label{learning-section}
For a fixed ordering of feature templates, to learn parameters that encourage the use of few feature templates, we look at the model as outputting not a single prediction but a sequence of prefix predictions $\{P_i\}$. Concretely, we optimize the following structured max-margin loss over training examples (with the dependence of $P$'s on $\mathbf{w}$ written explicitly where helpful):
\begin{align*}
\ell(x,y, \textbf{w}) = \sum_{k=1}^{i^*_y} h(P_k(x, \textbf{w}), y),~~~~~ \text{where}~~~~i^*_y = \min i ~~~~\text{s.t.}~~~~ h(P_i, y) = 0
\end{align*}
The per-example gradient of this objective for weights $\textbf{w}_j$ corresponding to feature template $\Phi_j$ then corresponds to
\begin{align*}
\frac{\partial \ell}{\partial \mathbf{w}_j} =&\sum_{k=j}^{i^*_y} \Phi_j(x, y_{\text{loss}}(P_k,y)) - \Phi_j(x, y)\\
\text{where}~~~~&y_{\text{loss}}(P_i,y) = \argmax_{y'} P_{i,y'} - m\cdot\mathds{1}(y' = y),
\end{align*}
where $\mathds{1}$ is an indicator function of the label $y$, used to define loss-augmented inference.

Since every prefix includes the prefix before it, we can see that for each training example, each feature template receives a number of hinge-loss gradients equal to its distance from the index where the margin requirement is finally reached. We add an $\ell_2$ regularization term to the objective, and tune the margin $m$ and the regularization strength to tradeoff between speed and accuracy.


We use a greedy stagewise approach to learn template ordering.
Given an ordered subset of templates, we add each remaining template
to our ordering and estimate parameters, selecting as the next
template the one that gives the highest increase in development set
performance.  

\section{Experimental Results}


We conduct our experiments on classifier-based, greedy part-of-speech tagging. Sequential classifiers achieve very strong performance on this task - for example, our classifier baseline achieves an accuracy of 97.22, while Stanford's CRF-based tagger scores 97.26. The high efficiency of baseline greedy POS tagging approaches makes this a particularly challenging domain in which to evaluate our speed-up algorithm. In contrast, it is easier to find gains in complex tasks with more overhead.

Our baseline greedy tagger uses the same features and factors described
described in \newcite{choi-palmer-12}. We evaluate our models on the Penn Treebank 
WSJ corpus \cite{marcus-et-al-93}, employing the typical POS train/test split. 
The parameters of our models are learned using AdaGrad \cite{Duchi2011} with $\ell_2$
regularization via regularized dual averaging \cite{Xiao2009}.

This baseline model ({\bf baseline}) tags at a rate of approximately 23,000 tokens per
second on a 2010 2.1GHz AMD Opteron machine with accuracy comparable to 
similar taggers \cite{Gimenez-Marquez-2004,choi-palmer-12,toutanova-et-al-03}.
On the same machine the greedy Stanford CoreNLP left3words
part-of-speech tagger also tags at approximately 23,000 tokens per
second. Significantly higher absolute speeds for all methods can be
attained on more modern machines.

We include additional baselines that
divide the features into templates, but train the templates'
parameters more simply than our algorithm. The {\bf stagewise} baseline
learns the model parameters
for each of the templates in order, starting with only one template---once each
template has been trained for a fixed number of iterations, that template's
parameters are fixed and we add the next one.
We also create a separately-trained baseline model for each fixed prefix of the feature templates ({\bf fixed}).  This shows that our speedups are not simply due to
superfluous features in the later templates.

Our main results are shown in Table~\ref{pos-compare-table}.
We increase the speed of our baseline POS tagger by a factor of 5.2x
without falling below 97\% test accuracy.  By tuning our training
method to more aggressively prune templates, we achieve speed-ups of over 10x
while providing accuracy higher than 96\%.

\begin{table}
\begin{center}
\begin{tabular}{|lllll|}
\hline
\bf Model/margin & \bf Tok. accuracy & \bf Unk. accuracy & \bf Feat. templates & \bf Speedup \\\hline\hline
Baseline & 97.22 & 88.63 & 46 & 1x\\ 
Stagewise & 96.54 & 83.63 & 9.50 & 2.74\\
\hline

Fixed & 89.88 & 56.25 & 1 & 16.16x\\ 
Fixed & 94.66 & 60.59 & 3 & 9.54x\\
Fixed & 96.16 & 87.09 & 5 & 7.02x\\ 
Fixed & 96.88 & 88.81 & 10 & 3.82x\\ \hline

Dynamic/15 & 96.09 & 83.12 & 1.92 & {\bf 10.36x}\\ 
Dynamic/35 & 97.02 & 88.26 & 4.33 & {\bf 5.22x}\\ 
Dynamic/45 & 97.16 & 88.84 & 5.87 & 3.97x\\ 
Dynamic/50 & {\bf 97.21} & 88.95 & 6.89 & 3.41x\\  \hline

\end{tabular}
\caption{Comparison of our models using different margins $m$, with speeds measured relative to the baseline. We train a model as accurate as the baseline while tagging 3.4x tokens/sec, and in another model maintain $>97\%$ accuracy while tagging 5.2x, and $>96\%$ accuracy with a speedup of 10.3x.\label{pos-compare-table}}
\vspace{-0.5cm}
\end{center}
\end{table}

\begin{figure}
\centering
\begin{subfigure}[b]{0.48\textwidth}
\includegraphics[width=\textwidth]{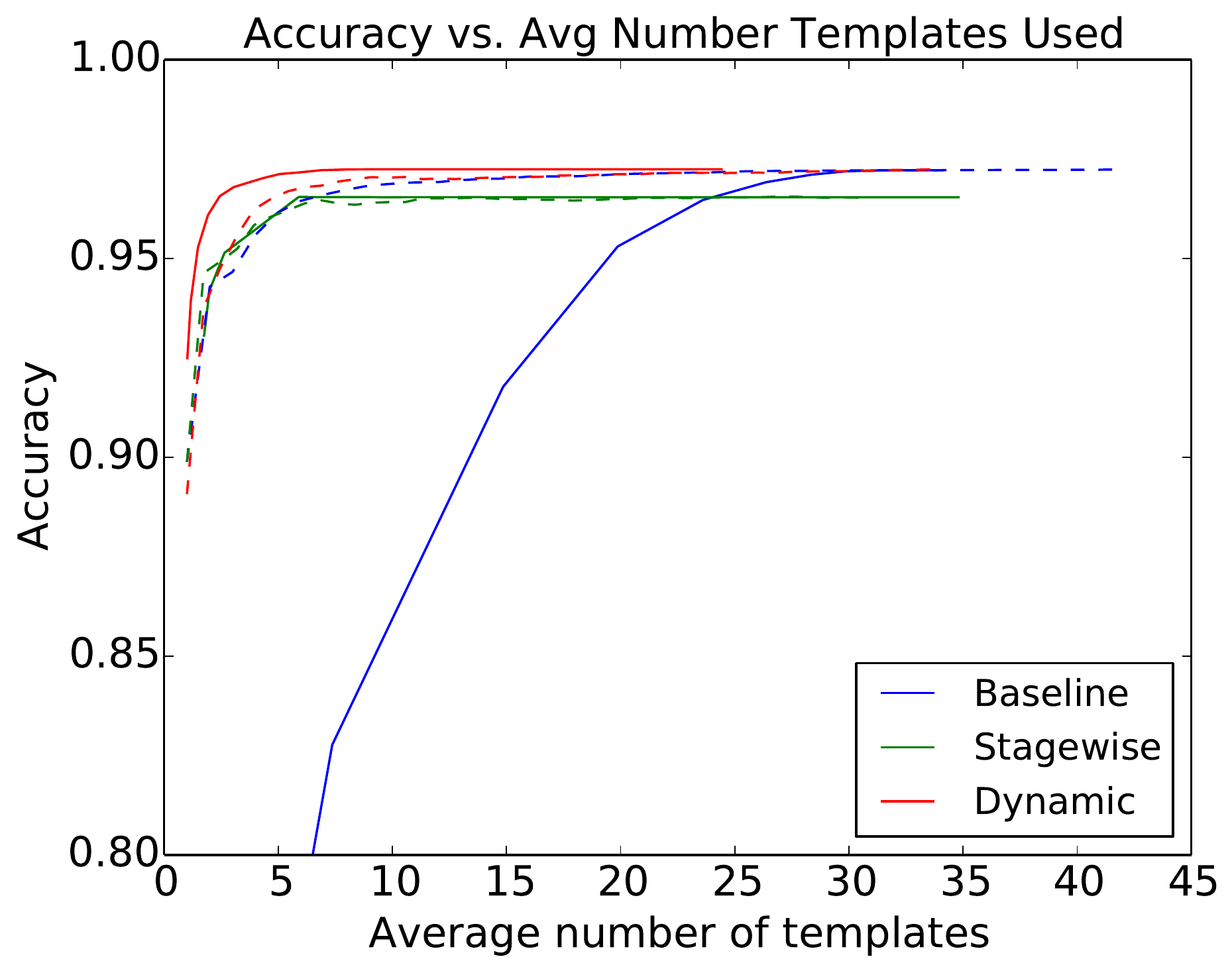}
\label{fig:tiger}
\end{subfigure}~
\begin{subfigure}[b]{0.5\textwidth}
\includegraphics[width=\textwidth]{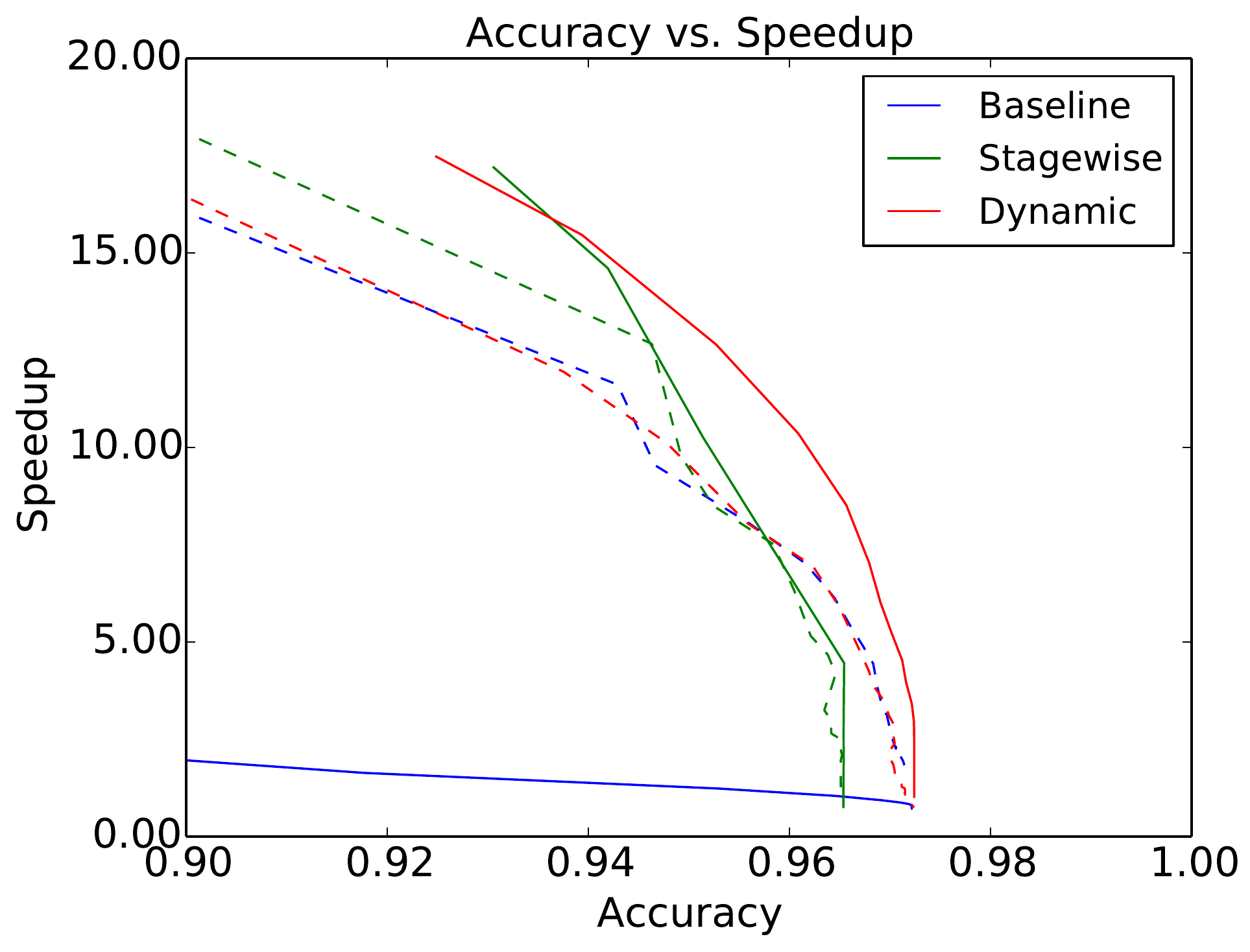}
\label{fig:mouse}
\end{subfigure}
\vspace{-0.5cm}
\caption{Left-hand plot depicts test accuracy as a function of the average number of templates used to predict. Right-hand plot shows speedup as a function of accuracy. In both plots, dashed lines correspond to varying the number of templates explicitly, whereas solid lines correspond to varying the margin. Dotted baselines are an ensemble of 46 models trained using each prefix of templates.}\label{fig:plots}
\vspace{-0.5cm}
\end{figure}

Results show our method ({\bf dynamic}) learns to dynamically select
the number of templates, often
using only a small fraction.
The majority of test tokens can be tagged using
only the first few templates: just over 40\% use one template, 
and 75\% require at most four
templates, while maintaining 97.17\% accuracy.
On average 6.71 out of 46 templates are used, though a small set of complicated
instances never surpass the margin and use all 46 templates. The right hand plot of Figure \ref{fig:plots} shows speedup vs. accuracy for various settings of the confidence margin $m$.

The left plot in Figure \ref{fig:plots}
depicts accuracy as a function of the number of templates used at
test time. We present results for both varying the number of templates directly (dashed) and margin (solid). The baseline model trained on all
templates performs very poorly when using margin-based inference, since its training objective does not learn to predict with only prefixes. When predicting using a fixed subset of templates, we use a different baseline model for each one of the 46 total template prefixes, learned with only those features; we then compare the test accuracy of
our dynamic model using template prefix $i$ to the baseline model trained on the fixed prefix $i$.
Our model performs just as well as these separately trained models,
demonstrating that our objective learns weights that allow
each prefix to act as its own high-quality classifier.

\subsubsection*{Acknowledgments}
This work was supported in part by the Center for Intelligent Information Retrieval, in part by DARPA under agreement number FA8750-13-2-0020, and in part by NSF grant \#CNS-0958392. The U.S. Government is authorized to reproduce and distribute reprint for Governmental purposes notwithstanding any copyright annotation thereon. Any opinions, findings and conclusions or recommendations expressed in this material are those of the authors and do not necessarily reflect those of the sponsor.

\bibliographystyle{unsrtnat}
\bibliography{nips2014}

\end{document}